\documentclass{article} 
\usepackage{iclr2023_conference,times}

\usepackage{amsmath,amsfonts,bm}

\def\eqref#1{equation~\ref{#1}}

\def\1{\bm{1}}

\DeclareMathAlphabet{\mathsfit}{\encodingdefault}{\sfdefault}{m}{sl}
\SetMathAlphabet{\mathsfit}{bold}{\encodingdefault}{\sfdefault}{bx}{n}

\usepackage[utf8]{inputenc}
\usepackage{graphicx}
\usepackage[labelfont=bf, font=footnotesize]{caption}
\usepackage{subcaption}

\usepackage{amsthm}
\usepackage{amsmath}
\usepackage{float}
\usepackage{xcolor}
\usepackage{nameref}
\usepackage{appendix}
\usepackage{algorithm}
\usepackage{algpseudocode}
    
\usepackage{etoolbox,lipsum}
\usepackage{multirow}

\theoremstyle{definition}

\author{José Pombal\textsuperscript{1,2,3}, Pedro Saleiro\textsuperscript{1}, Mário A. T. Figueiredo\textsuperscript{2,3} \& Pedro Bizarro\textsuperscript{1} \\
\textsuperscript{1}Feedzai \quad \textsuperscript{2}Instituto Superior Técnico, Universidade de Lisboa \quad \textsuperscript{3}Instituto de Telecomunicações \\
\texttt{pedro.saleiro@feedzai.com}
}

\title{Fairness-Aware Data Valuation \\
for Supervised Learning}

\iclrfinalcopy 

\begin{document}

\maketitle

\begin{abstract}
    \noindent
    %
    %
    %
    Data valuation is a ML field that studies the value of training instances towards a given predictive task. Although data bias is one of the main sources of downstream model unfairness, previous work in data valuation does not consider how training instances may influence both performance and fairness of ML models.
    Thus, we propose \textbf{F}airness-\textbf{A}ware \textbf{D}ata Valuati\textbf{O}n (FADO), a framework
    that can be used to incorporate fairness concerns into a series of ML-related tasks (e.g., data pre-processing, exploratory data analysis, active learning).
    %
    %
    We propose an entropy-based data valuation metric suited to address our two-pronged goal of maximizing both performance and fairness, which is more computationally efficient than existing metrics.
    We then show how FADO can be applied as the basis for unfairness mitigation pre-processing techniques.
    Our methods achieve promising results --- up to a 40 p.p. improvement in fairness at a less than 1 p.p. loss in performance compared to a baseline --- and promote fairness in a data-centric way, where a deeper understanding of data quality takes center stage.
\end{abstract}

\section{Introduction}\label{sec:intro}

A key ingredient in machine learning (ML) is the data used to train models \citep{hestnessDeepLearningScaling2017,najafabadiDeepLearningApplications2015}. 
Even specific instances have been shown to play a crucial part in the performance of ML models \citep{tonevaEmpiricalStudyExample2019,ngiamDomainAdaptiveTransfer2018,zhuLearningTransferLearn2020}: there can be more valuable and less valuable training data instances, with respect to a certain metric of interest.
The field of data valuation has arisen in this context, with the goal of associating each training datum with its corresponding value.
Several approaches for data valuation have been proposed \citep{ghorbaniDataShapleyEquitable2019,yoonDataValuationUsing2020,ghorbaniDistributionalFrameworkData2020a,kwonBetaShapleyUnified2022,simDataValuationMachine2022} and have been applied in a series of applications, such as noisy labels detection, and active learning, with varying degrees of success \citep{ghorbaniDataShapleyEquitable2019,yoonDataValuationUsing2020,kwonBetaShapleyUnified2022,ghorbaniDataShapleyValuation2021}.

However, along with their exceptional performance in some tasks, ML algorithms have been shown to have the potential to exacerbate and propagate biases embedded within data \citep{exacerbate-inequities-kirchner2016machine,exacerbate-inequities-howard2018ugly,exacerbate-inequities-o2016weapons}.
This is especially true if such models are trained with regards only to some performance metric, and if they are treated as unamenable black boxes.
Such a stance is particularly problematic in high-stakes decision-making domains, where biased policies may result in certain social groups being systematically discriminated, and locked out of important goods and services (e.g., by race, sex).
This phenomenon has sparked the creation of the field of fair ML, which has grown rapidly in recent years \citep{catonFairnessMachineLearning2020,mehrabiSurveyBiasFairness2021,pessachReviewFairnessMachine2022}.

Just as data plays a part in the performance of models, so does it play a part in their potential for discrimination~\citep{saleiro2020dealing,rodolfa2020bias,chakrabortyBiasMachineLearning2021,chenWhyMyClassifier2018}.
Indeed, data bias can shape the landscape of fairness and performance~\citep{pombalUnderstandingUnfairnessFraud2022,pombalPrisonersTheirOwn2022}, but little work has been made to understand the sources of these biases.
Thus, we propose fairness-aware data valuation (FADO), a data valuation framework that allows measuring which training instances contribute the most to both performance and fairness of models. 
We also propose an entropy-based data valuation metric that is specifically tailored to relate to model performance and data bias concerns.
%
%
We argue that the proposed framework embodies the ultimate goal of fair ML, which is to incorporate fairness concerns into ML systems, while keeping their performance high.
FADO can be used in the same ways as data valuation, and, as such, can be a vehicle to introduce fairness concerns into a series of upstream and downstream tasks, such as data pre-processing, active learning, data generation, and exploratory data analysis (EDA).
%

In summary, we make the following contributions:
\textbf{1)} we propose fairness-aware data valuation framework based on a notion of utility that incorporates the value of specific instances towards both performance and fairness (lack of data bias). It can be used with continuous and categorical targets and protected attributes, as well as to promote subgroup fairness (see Section~\ref{subsec:framework}).
\textbf{2)} we introduce an entropy-based notion of data value, which can be used to calculate an instance's value towards accurate prediction of a target $Y$, and data bias with respect to one or more protected attributes (see Section~\ref{sec:entropy}) in any predictive task.
\textbf{3)} we conduct an evaluation of how the framework may be leveraged for fairness-aware training data sampling and re-weighing, achieving promising results on a real world use-case.

\section{Background \& Related Work}\label{sec:bg}


This work focuses on supervised learning, where data valuation is usually framed in the following way~\citep{ghorbaniDataShapleyEquitable2019,ghorbaniDataShapleyValuation2021,ghorbaniDistributionalFrameworkData2020a,yoonDataValuationUsing2020,karlasDataDebuggingShapley2022}:
given a set of training data with \textit{n} data instances, each characterized by a feature vector \(x_i\) (in which one, or more, protected attribute(s) \(z_i\) may be included), and a target variable \(y_i\), the goal is to find the value \(v_i\) of each training instance, with respect to some value metric of interest \(V\), obtained by a learning algorithm \(f\) trained to predict \(y_i\), given \(x_i\).
How \(v_i\) is calculated varies from work to work.
\citet{ghorbaniDataShapleyEquitable2019} take a game-theoretical approach and calculate the Data Shapley value of each data instance.
%
%
Although this method enjoys a series of desirable properties, it is intractable to compute, and approximations have been proposed by the original authors and in other works \citep{kwonBetaShapleyUnified2022,ghorbaniDistributionalFrameworkData2020a, karlasDataDebuggingShapley2022}.
Conversely, \citet{yoonDataValuationUsing2020} take a reinforcement learning approach (DVRL), where a data valuator model is jointly trained with the predictor of the task from which \(V\) is calculated.

As for the metric of interest, \citet{ghorbaniDataShapleyEquitable2019} and \citet{yoonDataValuationUsing2020} set \textit{V} as the accuracy of a predictor \(f\) on unseen data (test set), with \(v_i\) being the importance of data instance \(i\) towards achieving the performance that predictor \(f\) obtains when trained on the whole dataset.
However, \(V\) can be any metric of interest; \citet{karlasDataDebuggingShapley2022}, for example, perform data valuation with equalized odds difference \citep{hardtEqualityOpportunitySupervised2016} as \(V\).
That said, a glaring gap in the literature is that no work defines \(V\) as a combination of performance and fairness.
These components, if fairness is mentioned, are always shown separately.
However, the main goal of fair ML is to maximize both predictive performance and fairness.
Furthermore, besides some applications in active learning and noisy instance detection, data valuation has never been used for fairness-promoting interventions.
To bridge this gap, we lay out a framework that defines \(V\) as a function of both performance and data bias (fairness), and show how it can be used to promote observational fairness.
Being an inherently data-centric task, we also believe that fairness-aware data valuation addresses the lack of exploratory data analysis tools in fair ML.

\section{Fairness-Aware Data Valuation}\label{sec:fado}

\subsection{Framework}\label{subsec:framework}

As mentioned before, the key component of our framework is the definition of \(V\) as a function of both performance and fairness, rather than having them be separate components.
Our task is not only to accurately predict \textit{Y}, but also to do it fairly.
We shall refer to our metric of interest as utility (\(U\)) from here on in.
This nomenclature is inspired by literature on the field of Economics, where the well-being (utility) of an individual is defined as a function of the utility derived from a weighted combination of factors (e.g., economic goods), rather than a single one.
In our case, utility is a function of performance and fairness; how it is maximized depends on the definition of the function and on the weights attributed to performance and fairness.
Thus, the utility of a single data instance can be written as \(U_i(v_{y_i}, v_{z_i})\), where \(v_{y_i}\) is the performance-related valuation of instance \(i\), and \(v_{z_i}\) is the fairness-related one.

%
%
%
We propose two types of utility functions; a linear scalarization one:

\begin{equation}
    U_i = \alpha v_{y_i} + (1 - \alpha) v_{z_i},
\end{equation}
where \(\alpha \in [0, 1]\) governs the relative weight placed on performance and fairness, and a multiplicative scalarization one, inspired by the Cobb-Douglas formulation~\citep{douglasCobbDouglasProductionFunction1976} (widely used in Economics):

\begin{equation}
    U_i = v_{y_i}^{\alpha} \cdot v_{z_i}^{1 - \alpha},
\end{equation}
where \(\alpha\) plays the same role as above. 
Each of these utilities abide by the principles of fair ML and, thus, the FADO framework, as they strike a balance between performance and fairness: both unfair, high-performing systems, and fair, low-performing systems, are inferior in terms of utility to systems that achieve acceptable values for both metrics.
However, the two functions encode distinct preferences as far as this balance goes.
For example, the multiplicative utility is zero if either valuation is zero, whereas the linear utility is not.
Thus, if a practitioner is not interested in gains in performance if they do not lead to gains in fairness, they should opt for the multiplicative utility.

This general framework allows reasoning about and obtaining fairness-aware valuations for data instances, which can then be used for downstream tasks.
Our framework can also be extended to applications concerned with subgroup fairness (fairness across a series of protected attributes and their combinations)
; one needs only to include several \(Z\) terms in the utility function.
For example, in a case where there are \textit{k} protected attributes \(Z_k\), a linear utility function can be defined as:

\begin{equation}
    U_i = \alpha v_{y_i} + \sum_{j=1}^{k} \beta_j v_{z_{j_i}}, 
\end{equation} 

%
In the next Section, we propose an entropy-based metric for FADO, which is versatile for both performance- and fairness-related data valuation.

\subsection{Entropy Metric}\label{sec:entropy}


The main challenge posed by the framework is to define and calculate a \(v_y\) for the predictive task at hand, and \(v_z\) for the fairness component of the utility.
These could be, for example, the Data Shapley value with respect to a validation set accuracy in the predictive task (\(v_y\)) and the Shapley value with respect to, for instance, a validation set equality of opportunity difference among protected groups in the same predictive task (\(v_z\)).
Such a choice would be appropriate within the framework, but entails two difficulties: first, it limits the user to a specific fairness metric, which as we have seen, often implies sacrificing others; second, data Shapley values are intractably hard to compute, thus they must be approximated.
Instead, we propose a more metric-agnostic valuation, which is inspired by the literature on uncertainty estimation and active learning (for \(v_y\)), and on the data bias taxonomy proposed by \citet{pombalUnderstandingUnfairnessFraud2022}.

For some data instance \(i\), we propose to define \(v_{y_i}\) as the corresponding prediction Shannon entropy, averaged over the output of one or more trained models\footnote{In the binary classification case; for multi-class, the metric could be entropy for non-binary distributions; for regression, the metric would have to be the prediction's difference from the sample mean, for example.}.
In binary classification, the Shannon entropy (\textit{E}) for uncertainty sampling for an instance \(i\) and prediction \(\hat{y}_i\) is:

\begin{equation}
    V_{y_i} = E_{y_i} = \hat{y}_i \cdot \log_2 \hat{y}_i + (1 - \hat{y}_i) \cdot \log_2 (1 - \hat{y}_i)
\end{equation}

The rationale behind choosing the entropy without the target label is inspired by active learning, where there is abundant data outside the training set, but a small budget to label it, and thus it is vital to choose which instances to label.
A popular method in active learning is \textit{uncertainty sampling} \citep{lewis1994sequential,settlesActiveLearningLiterature2009}, where an ML model trained on data subsequently chooses to obtain labels for the observations on which it is most unsure (with the highest entropy).
The idea is that the model will learn more from adding these observations to the training set, than from adding observations on which the model is sure, as the training data already contains that information.
This metric is not perfect, since being confident does not necessarily mean being right, and high entropy observations may simply be noisy and not necessarily useful.
However, it is a useful valuation heuristic, which has been shown to work well in many settings, including fraud detection~\citep{lewis1994sequential,yangBenchmarkComparisonActive2018,barataActiveLearningImbalanced2021}.
We extend this rationale to data valuation and attribute the highest values to the instances with the highest entropy. 
In other words, we attribute less value to instances where the model is very sure, since there is reason to believe that these encode redundant information to the task at hand. 

As for \(v_{z_i}\), the metric is the same, but using a predictor of \textit{Z}, rather than of \textit{Y}.
For an instance \(i\), a binary protected attribute \(z_i\), and predicted group \(\hat{z_i}\), we have:

\begin{equation}
    V_{z_i} = E_{z_i} = \hat{z}_i \cdot \log_2 \hat{z}_i + (1 - \hat{z})_i \cdot \log_2 (1 - \hat{z}_i)
\end{equation}

%
Notice that the metric is essentially the same for \(v_y\) and \(v_z\), but for different reasons.
In the first case, we want to prioritize observations on which the model had more difficulty during training, since these might contain less redundant information.
In the second case, where the variable in question (\textit{Z}) is not the target for the task at hand, and is seen by the model even at inference time, we prioritize observations where the model had more difficulty in establishing a relationship among \textit{X}, \textit{Y}, and \textit{Z}, leveraging the fact the model has no explicit incentive to draw such relationships.
This is directly related to mitigating the base bias condition of the taxonomy of \citet{pombalUnderstandingUnfairnessFraud2022} (\(\mathbb{P}[X, Y] \neq \mathbb{P}[X, Y | Z]\)), and so related to promoting fairness.
This metric can also be seen as a means to promote ``procedural'' fairness ~\citep{greenbergTaxonomyOrganizationalJustice1987} in the learning process of the model: we are ``telling'' the model to pay less attention during training to the protected attribute, and whatever variables may be related with it.

The resulting combination of \(v_y\) and \(v_z\) is a utility measure that balances both performance and fairness by assigning lower value to redundant information about \textit{Y} and excessive information about \textit{Z} in the data.
Using a linear scalarization utility function, we would get:

\begin{equation}
    U_i = \alpha E_{y_i} + (1 - \alpha) E_{z_{i}}
\end{equation}
%

Recently, \citet{xuOnlineDataValuation2022} proposed using entropy as a measure of data value for online ML tasks (with respect only to performance).
Other metrics, such as data shapley, are not suitable for these settings, as they require a fixed dataset.
Our rationale for using entropy is similar to that of Xu et al., but we extend it seamlessly to the context of data bias and fairness, and to offline ML tasks.
In terms of computation, our method requires only the additional training of a model for \textit{Z}.
This corresponds --- in our case of binary classification and a binary protected attribute --- to a roughly two-fold increase in training time, which is much faster than Data Shapley or DVRL alternatives.

\section{Unfairness Mitigation Pre-processing with FADO}\label{sec:applications}

\subsection{Dataset}\label{subsec:datasets}

Based on the proposed framework, we apply unfairness mitigation pre-processing interventions on AOF, a real-world large-scale bank account-opening fraud dataset. 
Fraud detection is an extremely pertinent field for fair ML, since holding a bank account is seen as a basic right in the European Union~\citep{basic_account_EU}.
In this use case of fraud, fraudsters attempt to impersonate someone to open an account, and quickly exhaust its line of credit.
Each row in the dataset corresponds to an application for opening a bank account, submitted via the online portal of a large retail bank.
Data was collected over an 8-month period and contains over 500K rows.
The first 6 months are used for training and the last 2 months are used for testing, mimicking the procedure of a real-world production environment\footnote{Because of client-related privacy issues, no further details on the data can be provided; however, the dataset introduced by \citet{jesus2022turning} was generated to be faithful to the one used here. Please see {https://www.kaggle.com/datasets/sgpjesus/bank-account-fraud-dataset-neurips-2022} for more information.}.
%
%

\subsection{Evaluation Framework}
Fraud rate (positive label prevalence) is about $ 1\% $ in both sets.
This means that a naïve classifier that labels all observations as \textit{not fraud} achieves a test set accuracy of almost $(99\%)$.
Such large class imbalance entails certain additional challenges for learning~\citep{heLearningImbalancedData2009}, and calls for a specific evaluation framework.
In particular, bank account providers are not willing to go above a certain level of FPR, because each false positive may lead to customer attrition (unhappy clients who may wish to leave the bank).
At an enterprise-wide level, this may represent losses that outweigh the gains of detecting fraud.
The goal is then to maximize the detection of fraudulent applicants (high global true positive rate, TPR), while maintaining low customer attrition (low global false positive rate).
As such, we evaluate the model's TPR at a fixed FPR, imposed as a business requirement in our case-study; we assess the FPR ceiling of $5\%$.
%
%
As for fairness, we consider three popular metrics for classification settings: the ratio of group-wise FPRs, \textit{predictive equality} \citep{corbett-daviesAlgorithmicDecisionMaking2017}, the ratio of group-wise FNRs, a form of \textit{equality of opportunity} \citep{hardtEqualityOpportunitySupervised2016}, and the ratio of group-wise predicted positive rate, \textit{demographic parity} \citep{dworkFairnessAwareness2012}.
The protected attribute used will be the client's age (to counteract \textit{ageism}).

%
%
%

\subsection{Experimental Setup}\label{subsec:experimental-setup}

We wish to apply some pre-processing method on our training data in order to maximize performance and fairness on unseen data.
%
%
%
To this end, our pre-processing method consists of three steps: first, computing each training instance's value with respect to \textit{Y} and \textit{Z}. 
Second, computing the utility of each instance, given a utility function and respective parameters. 
Third, sampling or reweighing the training set before training models.
When sampling is done, the instances with lower utility that belong to the protected group with lowest fraud prevalence are discarded first.
In the case of reweighting, the instances with higher utility are assigned larger weights during training.
We consider two strategies: utility-aware prevalence sampling (UASP), and utility-aware reweighting (UAR).
UASP consists of undersampling the data, such that protected groups end up with the same prevalence (fraud rate); in our dataset, this means discarding label negatives of the group with the lowest fraud rate, starting with the instances with the least utility, until fraud rates are balanced.
UAR implies assigning weights to each observation in the training set prior to training, where the weight corresponds to the instance's utility --- the full training set is used.
We experiment with several hyperparameter configurations of the above approaches (see Appendix \ref{app-fadv-hp}).

After each preprocessing intervention, the resulting dataset (and instance weights, if it is the case) is used to train 25 LighGBM models (with hyperparameters sampled from a grid), which are then used to make predictions on a test set (the same 25 models are used throughout).
LightGBM was chosen for being a state-of-the-art algorithm for tabular data, and for easily allowing weights to be assigned to instances during training.
%
%
To highlight the models that achieve the best performance-fairness trade-offs, we also analyze the Pareto frontier~\citep{pareto1919manuale} of the landscape.
We also outline the 80\% rule~\citep{80percentrule} threshold as a guideline for how fair models should be: the best models are those that achieve the highest performance above this threshold of fairness.
We compare our fairness-aware data valuation approaches to the case of no intervention (the full training set), random prevalence sampling (RPS, same as UASP, but observations are randomly discarded, instead of using a utility-based order) and reweighing (RW) \citep{kamiranDataPreprocessingTechniques2012}.

\subsection{Results \& Discussion}\label{subsec:proc}

\begin{figure*}[htpb]
    \centering
    \includegraphics[width=\linewidth]{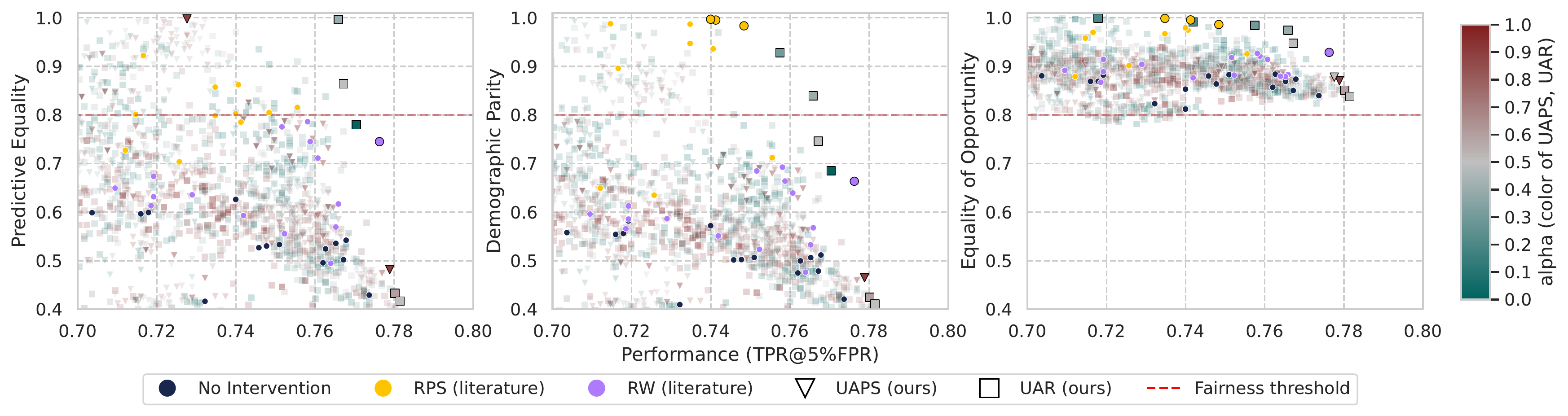}
    \caption{
Each point corresponds to one of 25 LightGBM models trained on the corresponding pre-processesed training set (see legend). 
    The axes represent performance and fairness on the test set in the task of predicting fraud (Y).
    Points that lie on the Pareto frontier  are represented by the larger opaque markers; small opaque markers are present to aid the visibility of baselines. 
    The \(\alpha\) colorbar represents the weight attributed to performance (\(\alpha\)) and fairness (1 - \(\alpha\)) in the utility function of our methods.
    %
    }
    \label{fig:dv-joint-zoom}
\end{figure*}

Figures \ref{fig:dv-joint-zoom} summarizes the performance-fairness trade-off landscape achieved by the models in the test set, trained on data that underwent the pre-processing techniques mentioned in the previous section.
It plots performance on the x-axis, and fairness on the y-axis; circles represent the no intervention and literature baselines, while triangles and squares represent our prevalence sampling and reweighing methods, respectively.
On our methods, the color varies between green and red.
Greener points were trained on data that was pre-processed with a higher weight on fairness in the FADO utility function; redder points had a higher weight placed on performance.
The underlying goal of this task is to maximize both performance and fairness, so the best points are the ones closest to the top right corner of the plots.
In particular, we care about the points with the highest performance above the 0.8 line, since this represents the aforementioned 80\% rule.

First, as expected, there seems to be a relationship between alpha (the parameter that governs the weight placed on performance and data bias in the utility functions) and the fairness-accuracy landscape --- the top performing models tend to be related to a higher \(\alpha\), while the fairest models feature lower \(\alpha\).
Furthermore, the best trade-offs on the Pareto frontier (high performance and high fairness) seem to be achieved by mid-range values (\(\alpha\) around 0.5), which suggests that a balanced valuation of points allows models to be much fairer at little loss in performance.
Notice, for example, on the leftmost plot of Figure~\ref{fig:dv-joint-zoom}, the opaque squares with TPR around 0.77 and fairness above 0.85.
It also seems that, for all fairness metrics, steep increases in fairness can be achieved with only slight drops in performance.
The relative abundance of points coloured between blue and red can be explained by the fact that for each of our proposed methods (UAPS, and UAR) there are more hyperparameters to choose from, and with which we experimented.

Comparing our methods with those from the literature, ours present the best performance above the 0.8 fairness line for all three fairness metrics tested.
Figure~\ref{fig:dv-joint-zoom} shows that even outside the Pareto frontier, our methods seem to occupy spaces of higher fairness, for a given level of performance, than RPS and RW --- especially our reweighing approach (UAR).
This meets our expectations, since our methods attempt to tackle all sources of data bias, instead of just prevalence disparity which is the case with RPS and RW (though the latter achieves this in an indirect way, without discarding examples).
Comparing among our own methods, reweighing approaches seem to obtain the best trade-offs. 
Not only do they feature several points on the Pareto frontier of all fairness metrics, but they also seem to perform better in general.
This was in part to be expected, as these methods do not require examples to be discarded, thus, in principle, leaving more information in the data for models to learn from.
Disregarding fairness for a moment, we can see that our approaches, for high values of \(\alpha\), are useful to increase the performance of ML models.
They can reach up to about 1 extra TPR point, when compared to the best performing models trained on the training set without intervention.

%
%



\section{Conclusions and Future Work}\label{sec:conclusions}

We introduced a framework and metric to relate performance and data bias to specific data instances.
Our framework can be applied with any type of classification or regression task, as well with any number of categorical and continuous protected attributes.
We also introduced an entropy-based notion of value and utility, which incorporates both performance and fairness concerns.
To validate the framework and the proposed metric, we showed how they could be applied in the context of an unfairness mitigation pre-processing intervention in a binary classification setting (training set sampling and reweighing).
A benchmark on our real-world fraud detection dataset showed that our methods often outperform existing ones in the literature, in terms of fairness-accuracy trade-offs.
In the future, we would like to extend this benchmark to inlcude other datasets and baselines.
We believe that tracing data bias to specific training instances is a promising practice to further the understanding of algorithmic unfairness and to add another layer of insight to data exploration and other tasks along the ML pipeline, like active learning --- a priority for future work as well.

\bibliography{refs}
\bibliographystyle{iclr2023_conference}

\vfill
\pagebreak

\appendix

\section{Algorithms to Compute Entropy}\label{subsec:entropy-algorithms}

Computing the required entropy-based metric for \(v\) is not as straightforward as in active learning, since our goal is to obtain valuations for a fully-labeled training set, and for two variables (Y and Z).
With this in mind, we propose two algorithms.
The first relies on creating bags from the training data of seen and unseen data, and training several models on each bag, from which entropy is obtained --- akin to k-fold cross-validation.
This is more closely related to active learning, since it involves a component of unseen data for which the model makes predictions.
The second algorithm is arguably more straightforward as it does not involve sampling the training set, but is more a measure of epistemic uncertainty rather than a combination of epistemic and aleatoric uncertainty~\citep{horaAleatoryEpistemicUncertainty1996} (as is the case with the active learning case)\footnote{Loosely speaking, the latter pertains to uncertainty incorporated by the model, while the former pertains to uncertainty inherent to the data.}.

\begin{algorithm}
    \caption{FADO Out-of-bag Entropy Algorithm}\label{alg:dv-bagging-based}
    \textbf{Input:} training set \textit{D}; variable of interest \textit{var}; number of bags \textit{n\_bags}; \% of train in each unseen set in each bag \textit{pct\_unseen}; models to train (pre-defined by user) \textit{M}\\
    \textbf{Output:} valuation vector with respect to \textit{var} \textit{V}
    \begin{algorithmic}[1]
        \State \textit{in\_bag\_sets}, \textit{out\_of\_bag\_sets} $\gets$ Split \textit{D} into \textit{n\_bags}, each with \textit{pct\_unseen} of \textit{D} into an out-of-bag set, and the rest into an in-bag set 
        \Comment{Data is sampled without replacement within each bag, but with replacement across bags; each observation must appear in at least one set of out-of-bag data}
        \State \textit{v\_list} \(\gets\) Empty list for intermediate target variable valuations
        \For{\texttt{in\_bag, out\_of\_bag \(\in\) [in\_bag\_sets, out\_of\_bag\_sets]}}
            \For{\texttt{m \(\in\) M}}
                \State \textit{fitted\_m} \(\gets\) Train \textit{m} on \textit{in\_bag} to predict \textit{var}
                \State \textit{v} \(\gets\) Vector of entropies of all predictions of \textit{fitted\_m} on \textit{out\_of\_bag}
                \State \textit{v\_list} \(\gets\) Append \textit{v}
            \EndFor
        \EndFor
        \State \textit{V} \(\gets\) Average entropies of each observation over all intermediate valuations in \textit{v\_list} 
    \end{algorithmic}
\end{algorithm}

\begin{algorithm}
    \caption{FADO In-bag Entropy Algorithm}\label{alg:dv-epistemic}
    \textbf{Input:} training set \textit{D}; variable of interest \textit{var}; models to train (pre-defined by user) \textit{M}\\
    \textbf{Output:} valuation vector with respect to \textit{var} \textit{V}
    \begin{algorithmic}[1]
        \For{\texttt{m \(\in\) M}}
            \State \textit{fitted\_m} \(\gets\) Train \textit{m} on \textit{D} to predict \textit{var}
            \State \textit{v} \(\gets\) Vector of entropies of all predictions of \textit{fitted\_m} on \textit{D}
            \State \textit{v\_list} \(\gets\) Append \textit{v}
        \EndFor
        \State \textit{V} \(\gets\) Average entropies of each observation over all intermediate valuations in \textit{v\_list} 
    \end{algorithmic}
\end{algorithm}

\begin{figure*}[h]
    \includegraphics[width=\linewidth]{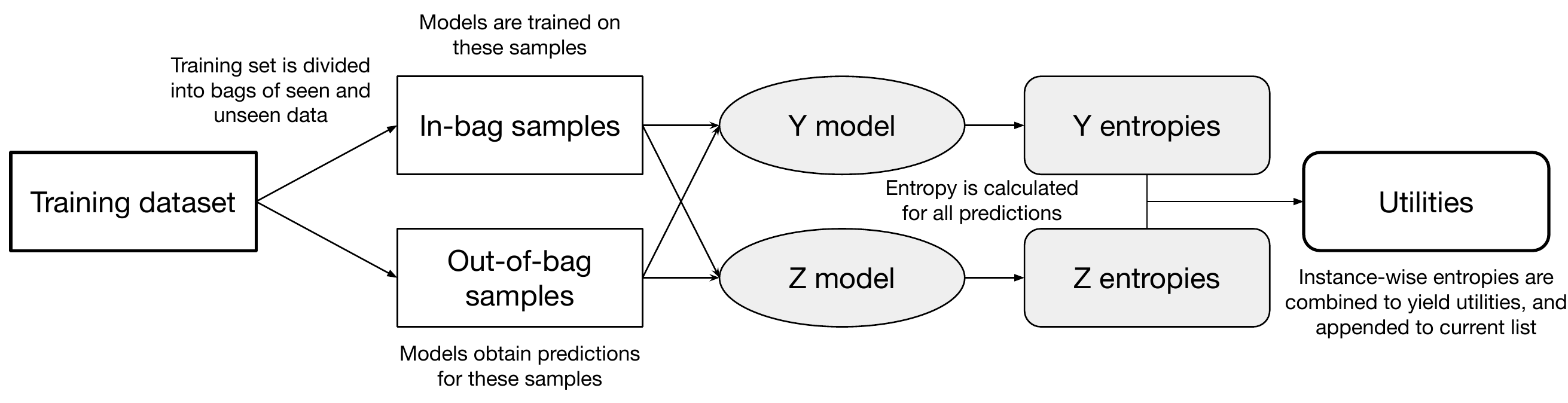}
    \caption{One iteration of the bagging-based algorithm, with one bag, one model for Y, and one model for Z.
    The final utility vector is obtained by averaging out all entropies obtained over each iteration, for each observation, and combining them into utilities.}
    \label{fig:dv-bagging-iteration}
\end{figure*}

To obtain a measure of utility, one must run either algorithm with Y as the target (to calculate \(v_y\)) and with Z as the target (to obtain \(v_z\)), for the same training set, models, and bagging splits (if algorithm 1 is chosen).
The final utility measure is then dependent on the chosen utility function, and its parameters.
As we argued in the previous section, we believe this entropy-based valuation is best suited for a utility function to which both \(v_y\) and \(v_z\) contribute positively (e.g., the linear and multiplicative utilities shown in Section~\ref{subsec:framework}).
In terms of the desiderata for data valuation strategies put forth recently by \citet{simDataValuationMachine2022}, our entropy-based metric satisfies D1-D3, and D7 and D9 if the sum of the entropies of each observation is seen to be the value of the whole dataset.

\section{Additional Results Plot}

\begin{figure*}[htpb]
    \centering
    \includegraphics[width=\linewidth]{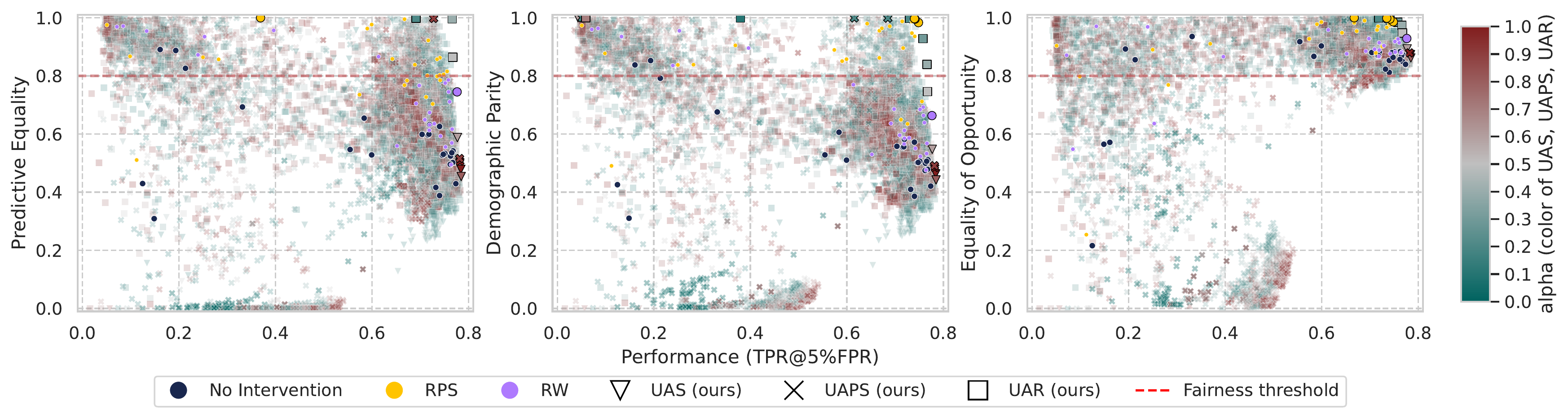}
    \caption{
    Zoomed out version of Figure~\ref{fig:dv-joint-zoom}.
    }
    \label{fig:dv-joint-full}
\end{figure*}

\section{Hyperparameters}
\label{app-fadv-hp}

\begin{table}[H]
    \centering
    \begin{tabular}{@{}llll@{}}
    \textit{\textbf{Hyperparameter\hspace{20pt}}} & \textit{\textbf{Distribution\hspace{20pt}}} & \textit{\textbf{Values\hspace{20pt}}} & \textit{\textbf{Description}} \\
    \textit{boosting\_type}      & -           & ``goss''        & Boosting type                \\
    \textit{n\_estimators}       & Log-uniform & \{20, 10000\} & Number of base tree learners \\
    \textit{num\_leaves}         & Log-uniform & \{10, 1000\}  & Maximum leaves of a tree     \\
    \textit{min\_child\_samples} & Log-uniform & \{5, 300\}    & Min. num. samples needed to create a leaf node                          \\
    \textit{max\_depth}          & Log-uniform & \{2, 20\}     & Maximum depth of a tree      \\
    \textit{learning\_rate}      & Log-uniform & {[}0.02, 0.5{]} & Boosting learning rate       \\
    \end{tabular}%
    \caption{Model hyperparameter grid for LightGBM.}
    \label{tab:app-hp-lgbm}
\end{table}

To compute data valuations, all permutations of the hyperparameters below were tried (\(v_y\) is the entropy of the Y classifier, \(v_z\) is the entropy of the Z classifier):

\textbf{Utility functions}: \(U = \alpha v_y + (1 - \alpha) v_z\), \(U = \alpha v_y - (1 - \alpha) v_z\),
\(U = v_y^{\alpha} v_z^{1 - \alpha}\);

\textbf{alpha} (\(\alpha\)): [0, 0.1, 0.2, 0.3, 0.4, 0.5, 0.6, 0.7, 0.8, 0.9, 1];


\textbf{Weights scaling}: ``min-max scaling'', ``no scaling'' (only used for UAR);

\textbf{Entropy algorithms}: ``bagging'', ``epistemic'' (the ones proposed in Section~\ref{subsec:entropy-algorithms});

\end{document}